\documentclass[preprint,review,12pt]{elsarticle}

\usepackage{amssymb}
\usepackage{amsmath}
\usepackage{booktabs}
\usepackage{floatrow}
\usepackage{textcomp}
\usepackage{mathtools}
\usepackage{physics}
\usepackage{appendix}
\usepackage{doi}
\usepackage{xcolor}

\journal{Journal of Sound and Vibration}

\begin{document}

\begin{frontmatter}



\title{Solving 2-D Helmholtz equation in the rectangular, circular, and elliptical domains using neural networks}


\author{D. Veerababu}
\author{Prasanta K. Ghosh\corref{cor1}}
\affiliation{organization={Department of Electrical Engineering},
            addressline={Indian Institute of Science}, 
            city={Bengaluru},
            postcode={560012}, 
            state={Karnataka},
            country={India}}

\cortext[cor1]{Corresponding author.}
\ead{prasantg@iisc.ac.in}

\begin{abstract}
Physics-informed neural networks offered an alternate way to solve several differential equations that govern complicated physics. However, their success in predicting the acoustic field is limited by the \emph{vanishing-gradient problem} that occurs when solving the Helmholtz equation. In this paper, a formulation is presented that addresses this difficulty. The problem of solving the two-dimensional Helmholtz equation with the prescribed boundary conditions is posed as an unconstrained optimization problem using \emph{trial solution method}. According to this method, a trial neural network that satisfies the given boundary conditions prior to the training process is constructed using the technique of transfinite interpolation and the theory of R-functions. This ansatz is initially applied to the rectangular domain and later extended to the circular and elliptical domains. The acoustic field predicted from the proposed formulation is compared with that obtained from the two-dimensional finite element methods. Good agreement is observed in all three domains considered. Minor limitations associated with the proposed formulation and their remedies are also discussed. 

\end{abstract}


\begin{highlights}
\item Two-dimensional Helmholtz equation is solved using physics-informed neural networks
\item Limitations of the Lagrange multiplier method are discussed
\item Theory of R-functions is successfully applied to physics-informed neural networks
\item Solutions to Bessel’s, and Mathieu’s equations are presented
\end{highlights}


\end{frontmatter}


\section{Introduction} \label{Sec:1} 
Neural networks are widely used in several applications, including image processing \cite{Belagali2020,Mannem2021}, natural language processing \cite{Kumar23,Tsiartas09}, speech recognition \cite{Abhay2021,Srinivasan2019}, autonomous vehicles \cite{Jing2021,Kuutti2021}, healthcare \cite{Shamshirband2021,Talukder2020}, and condition monitoring \cite{Rafiee2007,Wang2007,Janssens2016,Bangalore2017,Lai2021}, due to their ability to learn complex patterns from data. Researchers realized that neural networks are capable of learning not only from data but also from the differential equations that govern complicated physics. Earlier work in this area, now known as physics-informed neural networks (PINNs), was reported by Lagaris et al. \cite{Lagaris1998}, and McFall et al. \cite{McFall2009}. However, the progress of PINNs was very marginal because of unavailability of the computational resources at that time \cite{Cuomo2022}. Recent advances, especially in the architecture of GPUs, and the development of automatic differentiation algorithms \cite{Baydin2018}, played a pivotal role in the development of PINNs as alternative numerical tools to solve differential equations that govern complicated physical laws. Some of the governing equations that were attempted to solve using PINNs are as follows: Poisson equation \cite{Basir2022,Maddu2022}, Kirchhoff plate bending equation \cite{Muradova2021,Guo2019}, Advection-diffusion equation \cite{Wang2023,Calabro2021,Bihlo2024}, Euler-Bernoulli beam bending equation \cite{Kapoor2023,Bazmara2023}, Laplace equation \cite{Pan2024,Lim2022}, Eikonal equation \cite{Bin2021,Grubas2023,Chen2023}, Helmholtz equation \cite{Alkhalifah2021,Cui2022,Stanziola2021,Schmid2024}, Klein-Gordan equation \cite{Wang2021KG,Jagtap2020b,Shi2023}, Schr{\"o}dinger equation \cite{Raissi2019,Harcombe2023,Radu2022,Chuprov2024}, Allen-Cahn equation \cite{Mattey2022,Wight2021}, Navier-Stokes equations \cite{Jin2021,Amalinadhi2022,Oldenburg2022,Eivazi2022,Baymani2015}, Korteweg-de Vries equation \cite{Raissi2019,xiao2021,Zhou2024,Wen2023}, and Burger's equation \cite{Lim2022,Raissi2019,Savovic2023,Mathias2022}. This paper focuses on solving the two-dimensional Helmholtz equation using PINNs.

In general, the problem of solving a differential equation subject to given boundary conditions can be posed as a constraint optimization problem. This can be converted into an unconstrained optimization problem using the Lagrange multiplier $\lambda$ \cite{Lagaris1998,Basir2022,Maddu2022}. In other words, the constraints are satisfied in a \emph{soft} manner. Most of the governing equations mentioned in the above paragraph were solved using the Lagrange multiplier method. However, finding the appropriate $\lambda$ is not a trivial task, especially when solving the Helmholtz equation. This is attributed to the oscillatory nature of the solution that changes with frequency. As the frequency changes, the loss function associated with the Helmholtz equation also changes, making the chosen $\lambda$ obsolete for the updated frequency. To alleviate this problem, automatic $\lambda$ update algorithms have been developed \cite{Basir2022,Maddu2022,Van2022,Wang2021}. These algorithms dynamically update $\lambda$ with respect to iterations and balance the interplay between the loss functions (associated with the Helmholtz equation and the boundary conditions) during the optimization process. Most of these algorithms calculate the updated $\lambda$ value based on the statistical values of the loss gradients, especially the mean and standard deviation \cite{Maddu2022,Van2022,Wang2021}. Although these algorithms have proven to be successful, their applicability is limited by the manual tuning of the additional hyperparameters. The hyperparameters chosen for a particular range of frequencies may not be suitable for other frequencies. Therefore, there is a need to develop alternate frameworks that completely eliminate the cumbersome process of finding the Lagrange multipliers to solve the Helmholtz equation over a wide range of frequencies. 

Lagaris et al.~\cite{Lagaris1998} proposed a methodology in which a trial neural network is constructed in such a way that it always satisfies the boundary conditions before the training process. Since the trial neural network satisfies the boundary conditions, the corresponding loss functions and subsequently the Lagrange multipliers can be eliminated from the optimization procedure. This methodology is successfully demonstrated in one-dimensional problems \cite{Lagaris1998,Veerababu2024}, as the construction of functions that interpolate boundary values is relatively less complex in one-dimensional settings. Extension of this methodology to two-dimensional problems, especially the Helmholtz equation, is a challenging task. Sukumar et al.~\cite{Sukumar2022} demonstrated the possibility of constructing such interpolation functions using the theory of R-functions \cite{Shapiro1991}. R-functions are the functions whose \emph{signs} are solely determined by \emph{signs} of their arguments rather than their magnitudes \cite{Shapiro1991}. These functions are widely used in constructive solid geometry to create complex shapes for computer-aided design purposes \cite{Shapiro2007}. It is realized that by exploiting the Boolean nature of some of the primitive R-functions, it is possible to solve the boundary value problems that occur in mechanics \cite{Rvachev1995}. In this work, this ansatz has been extended to neural networks for solving the two-dimensional Helmholtz equation. The main contributions of this work are as follows:
\begin{enumerate}
    \item The limitations associated with the traditional Lagrange multiplier method while solving the Helmholtz equation at higher frequencies are demonstrated.
    \item Using the theory of R-functions in conjunction with the inverse distance weighting interpolation \cite{Lu2008}, the acoustic field is predicted in the
    \begin{enumerate}[(i)]
        \item Rectangular domain \cite{Deng1998,Hashimoto1989}
        \item Circular domain \cite{Jin2019}
        \item Elliptical domain \cite{Oliveira2014}
    \end{enumerate}
    Mathematically, (i), (ii), and (iii) are equivalent to solving the Helmholtz equation \cite{Morse1986}, Bessel's equation \cite{Bowman2003}, and Mathieu's equations \cite{McLachlan1947} using PINNs, respectively.
    \item Minor limitations associated with the proposed methodology are discussed in detail with the remedies.
\end{enumerate}

The article is organized as follows: Section~\ref{Sec:2} demonstrates a generalized PINNs formulation for the Helmholtz equation. Using the Lagrange multiplier method, the problem is converted into an unconstrained optimization problem and solved for the rectangular domain in Section~\ref{Sec:3}. The limitations of the method are also discussed in the same section. The proposed trial solution method is discussed in detail, and its applicability to other domain shapes is demonstrated in Section~\ref{Sec:4}. Minor limitations of the proposed methodology as well as the remedies are also discussed in the same section. The article is concluded in Section~\ref{Sec:5} with the final remarks.

\section{Neural network formulation for the Helmholtz equation} \label{Sec:2}
The acoustic field $\psi(\mathbf{x})$ in a given domain $\Omega$ can be obtained by solving the Helmholtz equation \cite{Morse1986}
\begin{equation}
    \left(\nabla^2+k^2\right)\psi(\mathbf{x}) = 0, \qquad \forall\,\, \mathbf{x}\in\Omega, \label{Eq:1}
\end{equation}
subjected to the boundary conditions 
\begin{equation}
    \psi(\mathbf{x}) = \psi_j(\mathbf{x}), \qquad \forall\,\,\mathbf{x}\in \partial\Omega_j, \label{Eq:2}
\end{equation}
where $k=2\pi f/c$ is the wavenumber, $f$ is the frequency, $c$ is the speed of sound, $\nabla^2$ is the Laplacian operator, and $\psi_j$ is the prescribed boundary value on the $j$-th boundary $\partial\Omega_j$, where $j=$ 1, 2, 3, ..., $M$. 

\begin{figure}[h!]
	\includegraphics[scale=0.9]{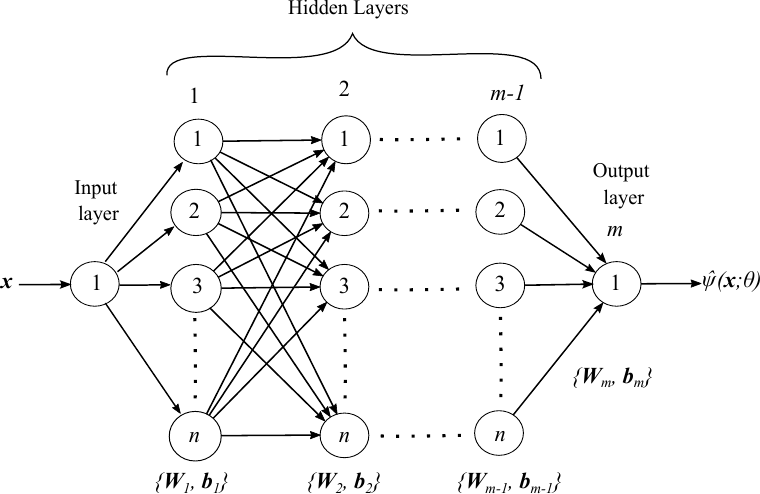}\centering
	\caption{\label{fig:1}{Schematic diagram of a feedforward neural network.}}
	\end{figure}

The \emph{universal approximation theorem} (UAT) states that a multilayered feedforward neural network with hidden layers containing a finite number of neurons can be used to approximate a continuous function, with an appropriate choice of activation function \cite{Hornik1989}. It implies that the acoustic field $\psi(\mathbf{x})$ can be approximated to a feedforward neural network $\hat{\psi}(\mathbf{x})$. However, the UAT does not provide sufficient information on the required number of hidden layers and neurons in them to capture the acoustic field. Usually, the acoustic field $\psi(\mathbf{x})$ is a nonlinear function of $\mathbf{x}$. Therefore, a feedforward neural network with multiple hidden layers is preferred over one with a single hidden layer to capture the nonlinear behavior of the acoustic field. A schematic diagram of a typical feedforward neural network that contains $m-1$ hidden layers with $n$ neurons in each layer is shown in Fig.~\ref{fig:1}. The set of $\{\mathbf{W}_q,\mathbf{b}_q\}$, where $q=$ 1, 2, 3,..., $m-1$ represent the weights and biases of $q$-th hidden layer, respectively.
 
 The feedforward neural network takes the domain information $\mathbf{x}$ in a discretized format at the input-layer. This information will be passed through the hidden layers where it will undergo a nonlinear activation function $\sigma$ along with the weights and biases $\{\mathbf{W}_q,\mathbf{b}_q\}$. The output from the hidden layers is then passed through the output-layer, where it undergoes a linear activation function unlike in the hidden layers. If $\mathbf{f}_0$, $\mathbf{f}_q$, and $\mathbf{f}_m$ represent the output of the input, hidden, and output-layers, respectively, then the network can be mathematically represented as follows
 \begin{align}
    \mathbf{f}_0 &= \mathbf{x}, \\
    \mathbf{f}_q &= \sigma(\mathbf{W}_q\mathbf{f}_{q-1}+\mathbf{b}_q), \qquad q=1, 2, 3,..., m-1, \\
    \mathbf{f}_m &= \mathbf{W}_m\mathbf{f}_{m-1}+\mathbf{b}_m.
\end{align}
 If $\hat{\psi}(\mathbf{x};\theta)$ represents the output of the neural network as shown in Fig.~\ref{fig:1}, then, $\hat{\psi}(\mathbf{x};\theta) = \mathbf{f}_m$. Here, $\theta = \{\mathbf{W}_q,\mathbf{b}_q, \mathbf{W}_m, \mathbf{b}_m\}$ are called the parameters of the network $\hat{\psi}(\mathbf{x};\theta)$. 

 The parameters $\theta$ can be obtained by solving the following optimization problem \cite{Basir2022}
 \begin{equation}
\begin{aligned}
\min_{\theta} \quad & \mathcal{L}_d(\mathbf{x}_d;\theta), \quad \mathbf{x}_d\in\Omega \\
\textrm{s.t.} \quad & \mathcal{L}_{b,j}(\mathbf{x}_{b,j};\theta)=0, \quad \mathbf{x}_{b,j}\in\partial\Omega_j \label{Eq:6}
\end{aligned}
\end{equation}
where $\mathcal{L}_d$ and $\mathcal{L}_{b,j}$ are the loss functions associated with the Helmholtz equation and the boundary conditions, respectively. They can be evaluated from Eqs.~(\ref{Eq:1}) and (\ref{Eq:2}), respectively, as follows
\begin{align}
    \mathcal{L}_d(\mathbf{x}_d;\theta) &= \frac{1}{N_d}\sum_{i=1}^{N_d}\left\|\left.\nabla^2\hat{\psi}(\mathbf{x};\theta)\right|_{\mathbf{x}=\mathbf{x}^{(i)}_d}+k^2\hat{\psi}(\mathbf{x}^{(i)}_d;\theta)\right\|^2_2 \label{Eq:7}, \\
    \mathcal{L}_{b,j}(\mathbf{x}_{b,j};\theta) &= \frac{1}{N_{b,j}}\sum_{i=1}^{N_{b,j}}\left\|\hat{\psi}(\mathbf{x}^{(i)}_{b,j};\theta)-\psi_j(\mathbf{x}^{(i)}_{b,j})\right\|^2_2. \label{Eq:8}
\end{align}
Here, $\left\|\,\cdot\,\right\|_2$ represents the $L_2$-norm, $N_d$ are the number of collocation points inside the domain, $N_{b,j}$ are the number of collocation points on the $j$-th boundary with $i$-th point being represented by $\mathbf{x}^{(i)}_d$ and $\mathbf{x}^{(i)}_{b,j}$, respectively.

Eq.~(\ref{Eq:6}) represents a constrained optimization problem. It can be solved by converting it into an unconstrained optimization problem using two methods: 1) Lagrange multiplier method, 2) Trial solution method. Note that the boundary conditions are referred to as constraints in this work. 

\section{Lagrange multiplier method} \label{Sec:3}
In this method, the loss functions associated with the boundary conditions ($\mathcal{L}_{b,j}$) are added to the loss function associated with the Helmholtz equation ($\mathcal{L}_d$) through the Lagrange multiplier $\lambda_j$, and the optimization is performed as follows \cite{Basir2022,Maddu2022,Raissi2019,Wang2021}
\begin{equation}
\min_{\theta} \quad \mathcal{L}_d+\sum_{j=1}^{M}\lambda_j\mathcal{L}_{b,j}. \label{Eq:9}
\end{equation}
It ensures that the boundary conditions given in Eq.~(\ref{Eq:2}) are satisfied in a \emph{soft} manner. 
	\begin{figure}[h!]
	\includegraphics[scale=0.9]{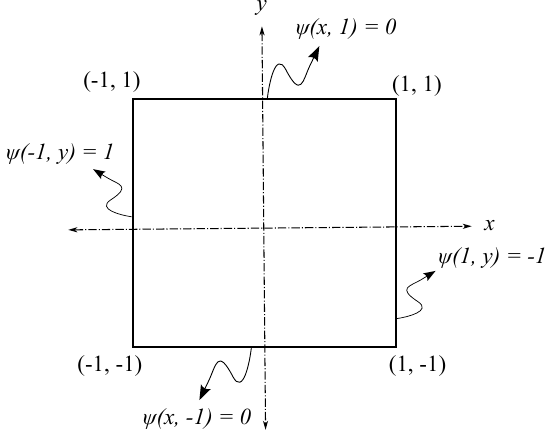}\centering
	\caption{\label{fig:2}{Schematic diagram of the rectangular domain with the boundary conditions.}}
	\end{figure}
 
The introduction of the Lagrange multiplier considerably reduces the complexity of the problem. However, finding the appropriate $\lambda_j$ is not a trivial task, especially for the acoustic problems described by Eqs.~(\ref{Eq:1}) and (\ref{Eq:2}). To demonstrate this, a rectangular domain of size 2 m $\times$ 2 m as shown in Fig.~\ref{fig:2} is chosen with the following boundary conditions
\begin{align}
    \psi(1,y) &= -1, \quad y\in [-1, 1] \\
    \psi(x,1) &= 0, \quad x\in [-1, 1] \\
    \psi(-1,y) &= 1, \quad y\in [-1, 1] \\
    \psi(x,-1) &= 0, \quad x\in [-1, 1]
\end{align}

\begin{figure}[h!]
	\includegraphics[scale=0.9]{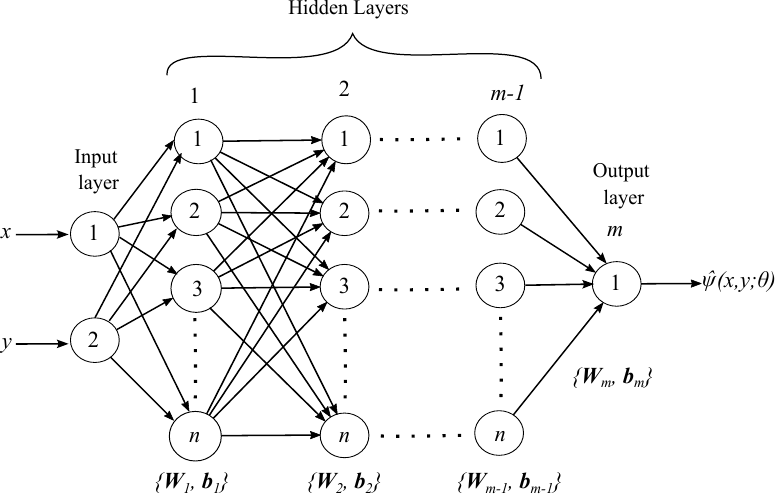}\centering
	\caption{\label{fig:3}{Schematic diagram of a feedforward neural network for the 2-D domain.}}
	\end{figure}

The acoustic field $\psi(x,y)$ can be obtained by solving the following two-dimensional (2-D) Helmholtz equation \cite{Morse1986}
\begin{equation} 
    \left(\frac{\partial^2}{\partial x^2}+\frac{\partial^2}{\partial y^2}+k^2\right)\psi(x,y) = 0. 
\end{equation}
To solve it, a feedforward neural network is constructed as shown in Fig.~\ref{fig:3}. It will have two neurons in the input-layer to feed the randomly discretized $x$ and $y$ coordinates into the network. 

\begin{table}[h!]
    \centering
    \begin{tabular}{@{}clc@{}}
        \toprule
        Sl. No. & \hfil Parameter & Value \\
        \midrule
        1 & No. of layers ($m$) & 7 \\
        2 & No. of neurons in each hidden-layer ($n$) & 90 \\
        3 & Activation function ($\sigma$) & $\sin$ \\
        4 & No. of internal collocation points ($N_d$) & 6400 \\
        5 & No. of boundary collocation points ($N_b$) & 320 \\
        6 & Optimizer & L-BFGS \\
        7 & No. of iterations & 20000 \\
        8 & Optimal tolerance & 10$^{-3}$ \\
        \bottomrule
    \end{tabular}
    \caption{Network architecture and training parameters.}
    \label{tab:1}
\end{table}

The network architecture and training parameters used for the training are presented in Table~\ref{tab:1}. It should be noted here that each boundary is divided into 80 collocation points, which add up to 320 for all four sides. Furthermore, the boundary points are uniformly distributed, unlike those of the internal domain, to cover the entire boundary with 5\% of the total internal collocation points. This discretization is carried out in similar lines with those reported in the literature \cite{Alkhalifah2021,Raissi2019,Wang2021}.

The loss function $\mathcal{L}_d$ is calculated as 
\begin{multline}
    \mathcal{L}_d(x_d,y_d;\theta) = \frac{1}{N_d}\sum_{i=1}^{N_d}\left\|\left.\left(\frac{\partial^2}{\partial x^2}\hat{\psi}(x,y^{(i)}_d;\theta)\right)\right|_{x=x_d^{(i)}}\right. \\
    +\left.\left(\frac{\partial^2}{\partial y^2}\hat{\psi}(x^{(i)}_d,y;\theta)\right)\right|_{y=y_d^{(i)}} \\
    +\left.k^2\hat{\psi}(x^{(i)}_d,y^{(i)}_d;\theta)\right\|^2_2, \label{Eq:15}
\end{multline}
and the loss functions $\mathcal{L}_{b,j}$ are calculated as 
\begin{equation}
    \mathcal{L}_{b,j}(x_{b,j},y_{b,j};\theta) = \frac{1}{N_{b,j}}\sum_{i=1}^{N_{b,j}}\left\|\hat{\psi}(x^{(i)}_{b,j},y^{(i)}_{b,j};\theta)-\psi_j(x^{(i)}_{b,j},y^{(i)}_{b,j})\right\|^2_2, \label{Eq:16}
\end{equation}
where $j=$ 1, 2, 3, and 4. For simplicity, consider the linear combination of all the individual boundary loss functions and denote the combined boundary loss function as $\mathcal{L}_b$, that is, 
\begin{equation}
    \mathcal{L}_{b}(x_b,y_b;\theta) = \frac{1}{N_{b}}\sum_{i=1}^{N_{b}}\left\|\hat{\psi}(x^{(i)}_{b},y^{(i)}_{b};\theta)-\psi(x^{(i)}_{b},y^{(i)}_{b})\right\|^2_2, \label{Eq:17}
\end{equation}

\begin{figure}[h!]
	\includegraphics[scale=0.85]{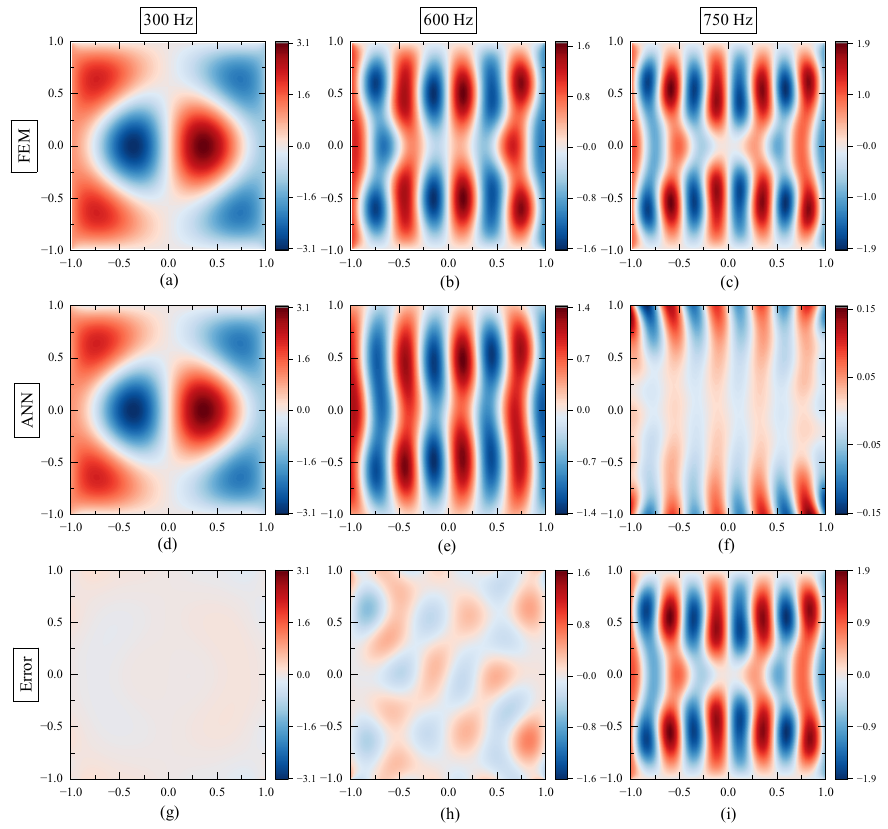}\centering
	\caption{\label{fig:4}{Acoustic field distribution at 300 Hz, 600 Hz, and 750 Hz: (a, b, c) from FEM, (d, e, f) from ANN ($\mathcal{L} = \mathcal{L}_d + \mathcal{L}_b$), and (g, h, i) are the errors (FEM-ANN).}}
	\end{figure}

According to Raissi et al. \cite{Raissi2019}, the Lagrange multiplier $\lambda_j$ is chosen as unity, that is, $\lambda = \lambda_j = 1$, and the overall loss function $\mathcal{L}$ is calculated as
\begin{equation}
    \mathcal{L} = \mathcal{L}_d + \lambda \mathcal{L}_b. 
\end{equation}
Using the training parameters mentioned in Table~\ref{tab:1}, an optimization procedure is performed to minimize the loss function $\mathcal{L}$ at frequencies $f=$ 300 Hz, 600 Hz, and 750 Hz with the speed of sound $c=$ 340 m/s. Implementation is carried out in MATLAB (Version R2022b) with the help of the Deep Learning Toolbox$^{\text{\texttrademark}}$ and the Statistics and Machine Learning Toolbox$^{\text{\texttrademark}}$. The acoustic field thus obtained from artificial neural networks (ANN) is compared with that obtained from the 2-D finite element model (FEM) developed using the COMSOL Multiphysics solver \cite{Comsol2022}, in Fig.~\ref{fig:4}. Refer to Appendix A for 2-D FEM modeling. 
 
 It can be observed from the results that the acoustic field obtained from the ANN is in good agreement with that obtained from the FEM at 300 Hz. As the frequency increases, that is, at 600 Hz, the agreement between the two methods deviates marginally. At a further higher frequency, that is, at 750 Hz, the predicted acoustic field completely deviates from that obtained from the FEM. This can be evidenced from the error plots drawn with the difference in the acoustic fields obtained from the two methods (FEM-ANN), in the last row of Fig.~\ref{fig:4}. 

\begin{figure}[h!]
	\includegraphics[scale=1.1]{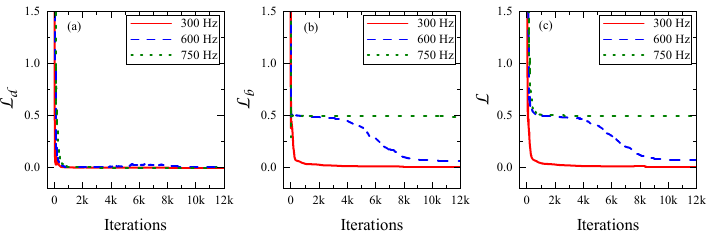}\centering
	\caption{\label{fig:5}{Individual loss functions at 300 Hz, 600 Hz, and 750 Hz: (a) $\mathcal{L}_d$, (b) $\mathcal{L}_b$, (c) $\mathcal{L}$.}}
	\end{figure}

 The reason for the deterioration of the predicted results with increasing frequency can be understood by observing the individual loss functions. Fig.~\ref{fig:5} shows the comparison of individual loss functions $\mathcal{L}_d$ and $\mathcal{L}_b$ with respect to iterations at 300 Hz, 600 Hz, and 750 Hz. It can be observed that the loss function associated with the Helmholtz equation $\mathcal{L}_d$ reduces to zero at all frequencies considered, while the loss function associated with the boundary conditions $\mathcal{L}_b$ does not decrease with increasing frequency. At 750 Hz, it stagnates around 0.5, which makes the overall loss function constant.  This ultimately stops the training process internally by halting the parameter update. Due to this, the network will not learn the underlying physics properly and the predicted results will be significantly different from the FEM results, as evidenced in Fig.~\ref{fig:4}. 

\begin{figure}[h!]
	\includegraphics[scale=1.2]{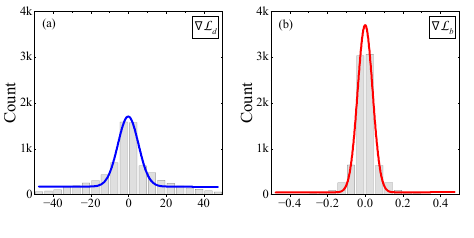}\centering
	\caption{\label{fig:6}{Gradients of the individual loss functions at 750 Hz: (a) $\nabla\mathcal{L}_d$, (b) $\nabla\mathcal{L}_b$.}}
	\end{figure}

 The reason for the bias of the training process at higher frequencies toward minimizing $\mathcal{L}_d$, leaving $\mathcal{L}_b$, can best be understood by comparing the individual loss gradients at 750 Hz. Fig.~\ref{fig:6} shows the histograms of the gradients of the individual loss functions with respect to the weights obtained at the last hidden layer and at the last iteration. Line graphs indicate the Gaussian distribution fits of the histogram bin values. Refer to Appendix B for the parameters of the distribution functions. 

 It can be observed from the histograms that the gradients of $\mathcal{L}_d$ are distributed over the range $[-40, 40]$. In contrast, the gradients of $\mathcal{L}_b$ are limited to $[-0.4, 0.4]$, and most of them are in the neighborhood of zero. This introduces a \emph{vanishing gradient problem} in the training process \cite{Wang2021}. In other words, the network stops minimizing $\mathcal{L}_b$ because most of its gradients are zero or close to zero, and continues to optimize $\mathcal{L}_d$. This problem can be bypassed by properly choosing the Lagrange multiplier $\lambda$, which is currently unity. Choosing it manually for each frequency is a cumbersome process. Therefore, automatic update algorithms have been developed in the past to dynamically update $\lambda$ based on the statistical values of the loss gradients \cite{Basir2022,Maddu2022,Van2022,Wang2021}. However, most of these algorithms involve manual tuning of the hyperparameters for each frequency, that is, the hyperparameter chosen for a particular frequency may not be suitable for other frequencies. For example, Eq. (\ref{Eq:18}) shows a popular algorithm that updates $\lambda$ using the standard deviation values of $\nabla\mathcal{L}_d$ and $\nabla\mathcal{L}_b$ \cite{Maddu2022}. 
\begin{equation}
    \lambda(\tau+1) = (1-\alpha)\lambda(\tau)+\alpha\hat{\lambda}(\tau), \label{Eq:18} 
\end{equation}
where
\begin{equation}
    \hat{\lambda}(\tau) = \frac{\text{max}\{\text{std}(\nabla\mathcal{L}_d(\tau)),\text{std}(\nabla\mathcal{L}_b(\tau))\}}{\text{std}(\nabla\mathcal{L}_b(\tau))}.
\end{equation}
Here, max$(\cdot)$ and std$(\cdot)$ denote the maximum value and standard deviation, respectively. The $\lambda$ at iteration $\tau+1$ is updated based on the value of $\lambda$ and $\hat{\lambda}$ at iteration $\tau$. Here, $\alpha$ is a hyperparameter that controls the contribution of $\lambda$ in the $\tau$-th iteration towards the $(\tau+1)$-th iteration. 
  \begin{figure}[h!]
    \includegraphics[scale=0.9]{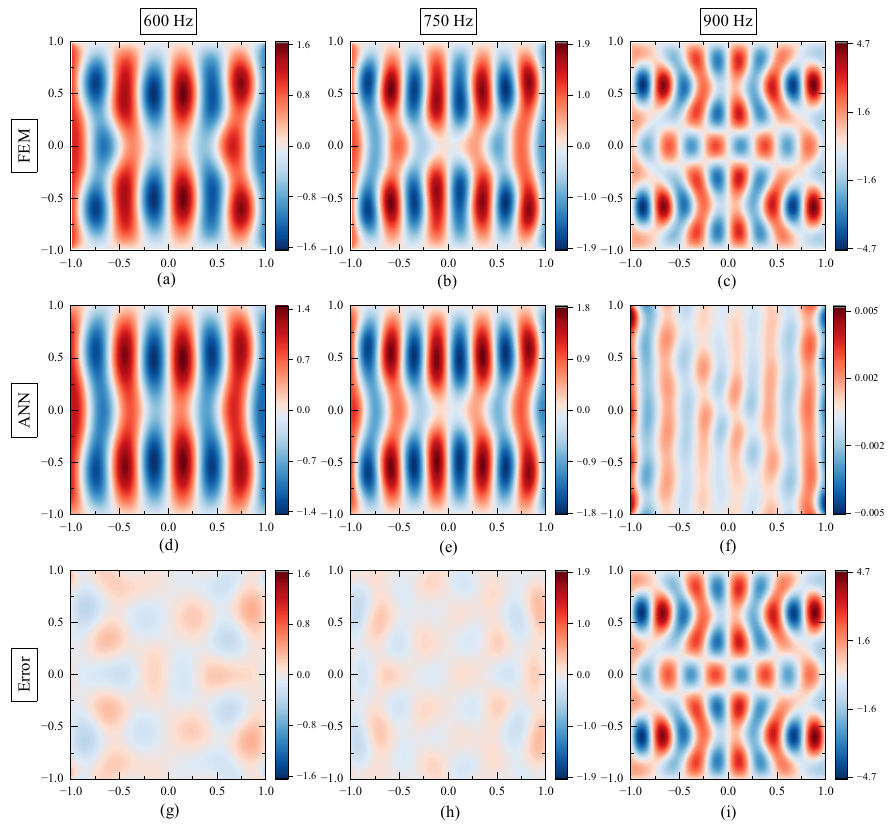}\centering
    \caption{\label{fig:7}{Acoustic field distribution at 300 Hz, 600 Hz, and 750 Hz: (a, b, c) from FEM, (d, e, f) from ANN (with dynamic $\lambda$), and (g, h, i) are the errors (FEM-ANN).}}
  \end{figure}  
  
Fig.~\ref{fig:7} shows the acoustic field distribution obtained using the $\lambda$ update rule mentioned in Eq.~(\ref{Eq:18}) with $\alpha=10^{-3}$ at frequencies 600 Hz, 750 Hz, and 900 Hz. It can be observed that with dynamically varying $\lambda$, the network is able to predict the acoustic field at 750 Hz with reasonable accuracy, unlike the one with $\lambda=$ 1 as shown in Fig.~\ref{fig:4}. However, the same update rule is not able to predict the acoustic field at 900 Hz. This is due to the fact that the chosen hyperparameter $\alpha$, which is suitable for 600 Hz and 750 Hz, is not suitable for 900 Hz. A suitable hyperparameter for 900 Hz has to be found using the \emph{trial-and-error} method. This implies that $\lambda$ update algorithms that require manual tuning of the hyperparameters are not suitable for solving the Helmholtz equation. Therefore, it is necessary to develop methodologies that bypass this difficulty and solve the Helmholtz equation at different frequencies without manual tuning of additional hyperparameters other than the network parameters mentioned in Table~\ref{tab:1}. In this article, an approach called \emph{trial solution method} is proposed that avoids the aforementioned difficulty and predicts the acoustic field in the frequency domain.  
     
\section{Trial solution method} \label{Sec:4}
In this method, a trial neural network $\hat{\psi}_t(\mathbf{x;\theta})$ will be constructed in such a way that it always satisfies the prescribed boundary conditions prior to the training process. This can be achieved by constructing $\hat{\psi}_t(\mathbf{x;\theta})$ in the form \cite{Lagaris1998,Sukumar2022}
\begin{equation}
      \hat{\psi}_t(\mathbf{x};\theta) = 
\begin{dcases}
    \psi_j(\mathbf{x}),& \mathbf{x}\in \partial\Omega_j \\
    \underbrace{\sum_{j=1}^{M}\phi_j(\mathbf{x})\psi_j(\mathbf{x}_{b,j})}_I+\underbrace{\vphantom{\sum_{j=1}^{M}} \phi_e(\mathbf{x})\hat{\psi}(\mathbf{x};\theta)}_{II}, & \mathbf{x}\in \Omega \label{Eq:20}
\end{dcases}
\end{equation}
where $\phi_j$ represents distance functions that interpolate the boundary conditions, and $\phi_e$ represents an equivalent distance function. It can be observed that the trial solution $\hat{\psi}_t(\mathbf{x};\theta)$ has two parts. The first part ensures that $\hat{\psi}_t(\mathbf{x};\theta)$ always satisfies the boundary conditions prior to the training process. It can be noticed that it does not contain any terms associated with the neural network $\hat{\psi}(\mathbf{x};\theta)$. Here, the distance function $\phi_j$ associated with the $j$-th boundary is constructed in such a way that its value is unity on the $j$-th boundary and is zero on the rest of the boundaries \cite{Lagaris1998,Sukumar2022}, that is, 
\begin{equation}
      \phi_j(\mathbf{x}) = 
\begin{dcases}
    1 \qquad \forall \,\mathbf{x}\in \partial\Omega_j, \\
    0 \qquad \forall \, \mathbf{x}\in \partial\Omega_{i\neq j}, \quad i = 1, 2, 3, ..., M.
\end{dcases}
\end{equation}
On the contrary, the second part contains the neural network $\hat{\psi}(\mathbf{x};\theta)$ and does not contain any terms associated with the boundary conditions. Here, the equivalent distance function $\phi_e$ is constructed in such a way that its value is zero on all the boundaries and is positive in the interior domain \cite{Lagaris1998,Sukumar2022}, that is,
\begin{equation}
      \phi_e(\mathbf{x}) = 
\begin{dcases}
    0 \qquad \forall \, \mathbf{x}\in \partial\Omega_j, \\
    >0 \qquad \forall \, \mathbf{x}\in \Omega. 
\end{dcases}
\end{equation}

  \begin{figure}[h!]
    \includegraphics[scale=1.0]{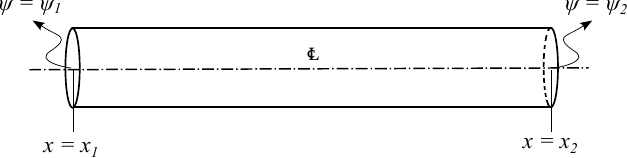}\centering
    \caption{\label{fig:8}{One-dimensional duct with the boundary conditions.}}
  \end{figure} 

In the case of the one-dimensional Helmholtz equation, the construction of the distance functions $\phi_j$ is a straightforward procedure. For example, the trial neural network $\hat{\psi}_t(x;\theta)$ that satisfies the boundary conditions of a one-dimensional (1-D) duct shown in Fig.~\ref{fig:8} can be constructed as follows
 \begin{equation}
    \hat{\psi}_t(x;\theta) = \frac{x_2-x}{x_2-x_1}\psi_1+\frac{x-x_1}{x_2-x_1}\psi_2+\left(\frac{x_2-x}{x_2-x_1}\right)\left(\frac{x-x_1}{x_2-x_1}\right)\hat{\psi}(x;\theta).  \label{Eq:23}
\end{equation}
Here, the distance functions $\phi_j$, $j=$ 1, 2, associated with the two boundaries are as follows
\begin{align}
    \phi_1 &= \frac{x-x_1}{x_2-x_1}, \\
    \phi_2 &= \frac{x_2-x}{x_2-x_1},
\end{align}
and the equivalent distance function $\phi_e$ is constructed as the product of the individual distance functions, that is, 
\begin{equation}
    \phi_e = \phi_1 \phi_2. \label{Eq:26}
\end{equation}
  \begin{figure}[h!]
    \includegraphics[scale=0.6]{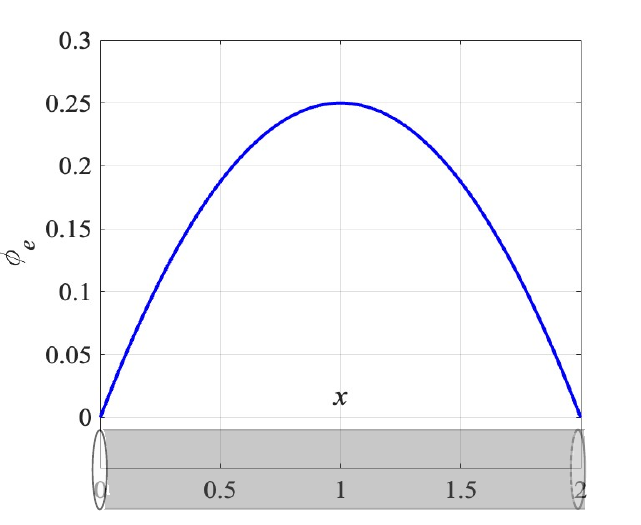}\centering
    \caption{\label{fig:9}{Equivalent distance function $\phi_e$ for the one-dimensional duct.}}
  \end{figure} 
  
Fig.~\ref{fig:9} shows the equivalent distance function $\phi_e$ for the 1-D duct of 2 units length, that is, $x_1=$ 0 and $x_2=$ 2. It can be seen that the function takes zero values at the boundary points and positive values at the interior points. Constructing a similar function in 2-D settings is not a trivial task. It is a problem of transfinite interpolation \cite{Rvachev2001}.

\subsection{Transfinite interpolation}\label{Sec:4.1}
It is a problem of constructing a surface that intersects a set of given curves. In other words, it is a process of constructing a single interpolation function that takes prescribed values on the boundaries \cite{Rvachev2001}. The problem of transfinite interpolation can be addressed using \emph{inverse distance weighting} interpolation \cite{Lu2008}, which is also known as Shepard's method. It is a popular interpolation technique that is used to interpolate scattered data obtained in the areas of geophysics, geology, meteorology, etc. \cite{Lu2008,Rvachev2001}. According to this method, the interpolation function $g(\mathbf{x})$ can be constructed as the weighted sum of the function values $g_i$ at locations $i=$ 1, 2, 3, ..., $p$. Mathematically, it is represented as
\begin{equation}
    g(\mathbf{x}) = \sum_{i=1}^{p}g_iW_i(\mathbf{x}). \label{Eq:27}
\end{equation}
Here, the weights $W_i$ are inversely proportional to the distance from the points ($\mathbf{x}_i$) where the function values $g_i$ are prescribed. 

The weight functions $W_i$ can be viewed as the basis functions using which the interpolation function $g$ is constructed. Therefore, it can be assumed that the weight functions are positive continuous functions that satisfy the interpolation condition \cite{Rvachev2001}
\begin{equation}
    W_i(\mathbf{x}_j) = \delta_{ij},
\end{equation}
and the partition of unity \cite{Rvachev2001}
\begin{equation}
    \sum_{i=1}^{p}W_i(\mathbf{x}) = 1, 
\end{equation}
where $\delta_{ij}$ is the Dirac delta function, and $j=$ 1, 2, 3, ..., $p$. 

Upon normalizing each inverse distance, the weight functions $W_i$ can be constructed as follows \cite{Rvachev2001}
\begin{equation} 
    W_i(\mathbf{x}) = \frac{\phi_i^{-1}(\mathbf{x})}{\sum_{j=1}^{p}\phi_j^{-1}(\mathbf{x})},
\end{equation}
where $\phi_i(\mathbf{x})$ is the distance between the point $\mathbf{x}$ and the point $\mathbf{x}_i$. Here, the weight functions $W_i$ become \emph{NaN} at $\mathbf{x} = \mathbf{x}_i$. Hence, its modified form, suggested by Rvachev et al. \cite{Rvachev2001} is used throughout this article and is presented below
\begin{equation}
    W_i(\mathbf{x}) = \frac{\Pi_{j=1;j\neq i}^{p}\phi_j(\mathbf{x})}{\sum_{k=1}^{p}\Pi_{j=1;j\neq k}^{p}\phi_j(\mathbf{x})}. \label{Eq:31}
\end{equation}

In addition to the interpolation property, the distance functions $\phi_i$ should have a differentiable property that is essential for the neural network training procedure. This is guaranteed if the distance functions that represent a geometric object (point set) can be expressed in the implicit function form $\phi_i(\mathbf{x})\geq$ 0 \cite{Rvachev2001}. It is known that implicit functions are often encountered while solving problems in the area of structural dynamics, fluid dynamics, electrical circuits, chemical kinetics, etc. \cite{Komornik2017}. On the other hand, the implicit representation of geometric objects like a line, a circle, an ellipse, etc., finds extensive applications in image processing, computer graphics, robotics, etc. \cite{Komornik2017}. Implicit functions with guaranteed differential properties can be constructed systematically using the theory of R-functions \cite{Shapiro1991}.

\subsubsection{R-functions}\label{Sec:4.1.1}
R-functions are the real-valued functions whose sign is completely determined by the sign of its arguments \cite{Shapiro1991}. For example, $\omega_1=xyz$ is a function whose sign depends only on the sign of its arguments $x$, $y$ and $z$. Therefore, it can be considered as an R-function. In contrast, $\omega_2=xyz+1$ is not an R-function because its sign depends not only on the signs of $x$, $y$ and $z$, but also on their magnitudes. Formally, if $\mathcal{S}$ represents the sign of a real-valued function $\Upsilon(\omega_1,\omega_2,\omega_3, ..., \omega_p)$, where $\omega_i$, $i=$ 1, 2, 3, ..., $p$ are the R-functions, then $\Upsilon$ is an R-function if \cite{Shapiro1991}
\begin{equation}
    \mathcal{S}[\Upsilon(\omega_1,\omega_2,\omega_3, ..., \omega_p)] = \Upsilon[(\mathcal{S}(\omega_1),\mathcal{S}(\omega_2),\mathcal{S}(\omega_3), ..., \mathcal{S}(\omega_p))].
\end{equation}
Note here that the R-functions are represented by $\omega_i$ to maintain consistency with the notation used by Rvachev in his work \cite{Rvachev1995}. Throughout the article, $\omega_i$ denotes the R-functions (not the angular frequencies) unless otherwise stated. 

The R-functions are extensively used in constructive solid geometry to create complex shapes \cite{Shapiro2007}. The required distance functions $\phi_i$ in Eq.~(\ref{Eq:31}) can be constructed by exploiting the sign property of the R-functions, that is, by categorizing the R-functions based on the values they take inside the domain and outside it. The expression for the distance function of a line segment joining the two points $\mathbf{x}_1=(x_1,y_1)$ and $\mathbf{x}_2=(x_2,y_2)$ which is constructed using the theory of R-functions is as follows \cite{Rvachev2001}
\begin{equation}
    \phi(\mathbf{x}) = \sqrt{h^2+\left(\frac{\sqrt{t^2+h^4}-t}{2}\right)^2}, \label{Eq:33}
\end{equation}
where $h$ is the signed distance function defined as 
\begin{equation}
    h(\mathbf{x})\coloneqq \frac{(x-x_1)(y_2-y_1)-(y-y_1)(x_2-x_1)}{L}, \label{Eq:34}
\end{equation}
and $t$ is the trimming function which is used to truncate the real axis at the points $\mathbf{x}_1$ and $\mathbf{x}_2$. In general, a normalized circular disk of radius $L/2$ can be used as a trimming function for a line segment \cite{Shapiro1999}, that is, 
\begin{equation}
    t(\mathbf{x}) \coloneqq \frac{1}{L}\left[\left(\frac{L}{2}\right)^2-\left\|\mathbf{x}-\mathbf{x}_c\right\|^2\right]. \label{Eq:35}
\end{equation}
Here, $L=\left\|\mathbf{x}_2-\mathbf{x}_1\right\|$ is the length of the line segment, and $\mathbf{x}_c=(\mathbf{x}_1+\mathbf{x}_2)/2$ is the center of the line segment.  
  \begin{figure}[h!]
    \includegraphics[scale=1.6]{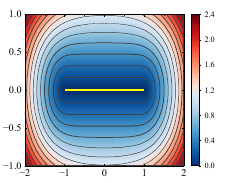}\centering
    \caption{\label{fig:10}{Distance function $\phi$ for the line segment joining the points $\mathbf{x}_1=(-1,0)$ and $\mathbf{x}_2=(1,0)$.}}
  \end{figure}
  
The distance function $\phi(\mathbf{x})$ in Eq.~(\ref{Eq:33}) is plotted for the line segment joining the points $\mathbf{x}_1=(-1,0)$ and $\mathbf{x}_2=(1,0)$ in the domain $x\in [-2, 2]$ and $y\in [-1, 1]$, in Fig.~\ref{fig:10}. It can be seen that the value of the function $\phi(\mathbf{x})$ decreases as $\mathbf{x}=(x,y)$ approaches the line segment. This is due to the fact that the value of the signed distance function $h(\mathbf{x})$ vanishes at any point on the line segment joining the points $\mathbf{x}_1$ and $\mathbf{x}_2$. 
 
\subsubsection{Transfinite interpolation using the R-functions}\label{Sec:4.1.2}
Recall that the main purpose of studying R-functions is to solve the transfinite interpolation problem, that is, to construct a surface that takes prescribed values on the boundaries. This can be done by categorizing the R-functions as positive and negative functions based on the values they take inside the domain and outside it, respectively. Upon categorization, R-functions can be treated as Boolean functions; positive values can be assigned to logical \emph{true} and negative values can be assigned to logical \emph{false} \cite{Shapiro2007}. Since Boolean functions are closed under composition, the R-function that represents the required surface can be constructed from the primitive R-functions $\omega_i$. The simplest R-functions that can be constructed from the set theory of Boolean functions are $\max(\omega_1,\omega_2)$ and $\min(\omega_1,\omega_2)$ using the disjunction $(\omega_1\land \omega_2)$ and the conjuction $(\omega_1\lor \omega_2)$ properties, respectively. The logical functions $\max(\omega_1,\omega_2)$ and $\min(\omega_1,\omega_2)$ can be expressed in terms of the algebraic R-functions as follows \cite{Shapiro1991,Rvachev2001}
\begin{align}
    \max(\omega_1,\omega_2) = \omega_1\land \omega_2 &= \frac{1}{2}(\omega_1+\omega_2+\abs{\omega_1-\omega_2}), \\
    \min(\omega_1,\omega_2) = \omega_1\lor \omega_2 &= \frac{1}{2}(\omega_1+\omega_2-\abs{\omega_1-\omega_2}),
\end{align}
where $\abs{\cdot}$ represents the absolute value of the argument. Generalized forms of these functions can be written as \cite{Shapiro2007}
\begin{align}
    \omega_1\land_{\beta} \omega_2 &= \frac{1}{1+\beta}\left(\omega_1+\omega_2+\sqrt{\omega_1^2+\omega_2^2-2\beta \omega_1\omega_2}\right), \label{Eq:38} \\ 
    \omega_1\lor_{\beta} \omega_2 &= \frac{1}{1+\beta}\left(\omega_1+\omega_2-\sqrt{\omega_1^2+\omega_2^2-2\beta \omega_1\omega_2}\right), \label{Eq:39}
\end{align}
where $\beta$ is an arbitrary parameter that takes the values between $(-1, 1)$. Through this generalization, properties other than \texttt{max} and \texttt{min} can be established. For example, if $\omega_1$ and $\omega_2$ represent the sides of a triangle, then Eqs.~(\ref{Eq:38}) and (\ref{Eq:39}) establish the triangle inequality \cite{Sukumar2022}. The property of differentiability is essential to use these functions in conjunction with neural networks. The functions in Eqs.~(\ref{Eq:38}) and (\ref{Eq:39}) can be made $s$ times differentiable by modifying them as \cite{Biswas2004}
\begin{align}
    \omega_1\land_{\beta}^{s} \omega_2 &= (\omega_1\land_{\beta} \omega_2)(\omega_1^2+\omega_2^2)^{s/2}, \label{Eq:41} \\ 
    \omega_1\lor_{\beta}^{s} \omega_2 &= (\omega_1\lor_{\beta} \omega_2)(\omega_1^2+\omega_2^2)^{s/2}. \label{Eq:42} 
\end{align}

Using this concept, the interpolation functions for rectangular, circular, and elliptical geometries were evaluated, and the acoustic fields were predicted in the subsequent sections. 
    
\subsection{Acoustic field in a rectangular domain} \label{Sec:4.2}
\begin{figure}[h!]
	\includegraphics[scale=0.9]{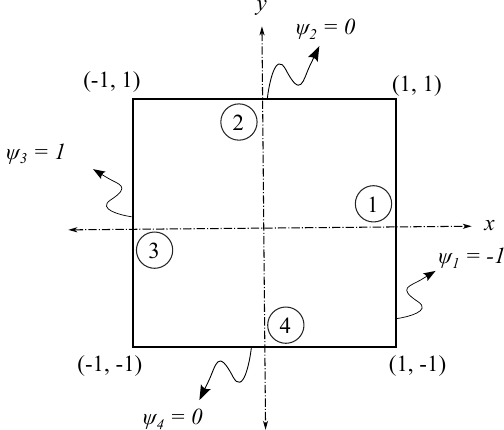}\centering
	\caption{\label{fig:11}{Schematic diagram of the rectangular domain with notation.}}
    \end{figure}

Let us consider the rectangular domain shown in Fig.~\ref{fig:2}. Name the sides and boundary values with numerals in the counterclockwise direction, as shown in Fig.~\ref{fig:11}. The domain consists of four line segments that join the points $(1, -1)$, $(1, 1)$, $(-1, 1)$, and $(-1, -1)$. 
        \begin{figure}[h!]
	\includegraphics[scale=1.7]{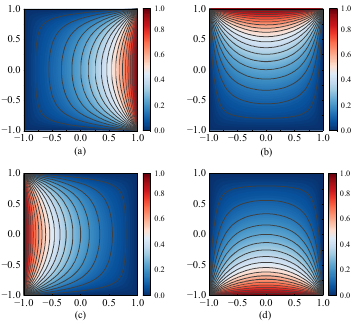}\centering
	\caption{\label{fig:12}{Weight functions for all the sides of the rectangular domain: (a) Side-1, (b) Side-2, (c) Side-3, (d) Side-4.}}
    \end{figure}
Using the theory of R-functions and the inverse distance weighting interpolation (Eq.~(\ref{Eq:27})), the first part of the trial solution in Eq.~(\ref{Eq:20}) can be constructed as follows
\begin{equation}
    \hat{\psi}_{t,I} = \sum_{i=1}^{4}W_i \psi_i, \label{Eq:44}
\end{equation}
where 
\begin{equation}
    W_i(\mathbf{x}) = \frac{\Pi_{j=1;j\neq i}^{4}\phi_j(\mathbf{x})}{\sum_{k=1}^{4}\Pi_{j=1;j\neq k}^{4}\phi_j(\mathbf{x})}. \label{Eq:45}
\end{equation}
Here, the distance function $\phi_j$ for each line segment is evaluated using Eq.~(\ref{Eq:33}). The weight functions thus obtained are shown in Fig.~\ref{fig:12}. It can be observed that the weight function of a particular side takes the value of unity on that side and is zero on the rest of the sides.

Now, the second part of the trial solution, that is, 
\begin{equation}
    \hat{\psi}_{t,II} = \phi_e  \hat{\psi} \label{Eq:46}
\end{equation}
can be obtained after careful evaluation of the equivalent distance function $\phi_e$. As discussed earlier, the equivalent distance function should vanish on the boundaries and should take a positive value inside the domain. The latter condition is to avoid the singularity of the distance function inside the domain \cite{Sukumar2022}. One of the simplest ways through which $\phi_e$ can be constructed is by multiplying individual distance functions as mentioned in the 1-D case in Eq.~(\ref{Eq:26})
\begin{equation}
    \phi_e = \prod_{i=1}^4\phi_i.  \label{Eq:47}
\end{equation}
However, the equivalent distance function thus obtained cannot be guaranteed to be differentiable up to the order $s$. An alternative way is to construct it using the theory of R-functions. 

Suppose that the chosen rectangular domain $\Omega$ is composed of four boundaries $\partial\Omega_i$, where $i=$ 1, 2, 3, and 4, and is represented by the following set operations
\begin{equation}
    \Omega = \partial\Omega_1 \cap \partial\Omega_2 \cap \partial\Omega_3 \cap \partial\Omega_4. \label{Eq:48}
\end{equation}
Let us say that each subdomain $\partial\Omega_i$ is represented by the primitive implicit R-function $\omega_i\geq 0$, then the R-function $\omega$ that represents the entire domain $\Omega$ can be found by replacing the symbol $\cap$ with  $\land$ in Eq.~(\ref{Eq:48}), that is, 
\begin{equation}
    \omega = \omega_1 \land_0^s \omega_2 \land_0^s \omega_3 \land_0^s \omega_4,
\end{equation}
where $\beta$ in Eqs.~(\ref{Eq:38}) is considered as zero. Here, $\omega$ is the required transfinite interpolation function which can be calculated using Eq.~(\ref{Eq:41}). Using this theory, $\phi_e$ can be constructed using the disjunction properties of the set theory as follows
\begin{equation}
    \phi_e = \phi_1 \land_0^s \phi_2 \land_0^s \phi_3 \land_0^s \phi_4.
\end{equation}
Even though this construction attains the differentiable property, it suffers from the lack of the associative property. Therefore, in this work a modified version of Eq.~(\ref{Eq:47}) proposed by  Biswas et al. \cite{Biswas2004} and Rvachev \cite{Rvachev1995}, which has all the desirable properties mentioned earlier, is used and is presented below
\begin{equation}
    \phi_e = \frac{1}{\sqrt[s]{\sum_{i=1}^4\frac{1}{\phi_i^s}}}.
\end{equation}
    \begin{figure}[h!]
	\includegraphics[scale=1.7]{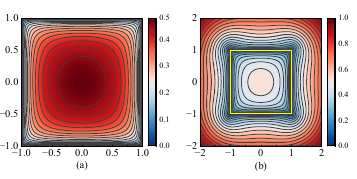}\centering
	\caption{\label{fig:13}{Equivalent distance function for the rectangular domain: (a) 2 m $\times$ 2 m, (b) 4 m $\times$ 4 m.}}
    \end{figure}
    
Fig.~\ref{fig:13} shows the equivalent distance function $\phi_e$ for the rectangular domain considered. It can be observed from Fig.~\ref{fig:13}a that $\phi_e$ vanishes at the boundaries and is positive inside the domain. This can be made more apparent by calculating $\phi_e$ for points outside the 2 m $\times$ 2 m domain  along with the domain and boundary points. Figure~\ref{fig:13}b shows the equivalent distance function $\phi_e$ plotted over the 4 m $\times$ 4 m domain. It can be observed that $\phi_e$ is positive inside the domain and vanishes on the boundary. A yellow reference boarder is drawn to demarcate the domain and its complement at the boundary.

Now, the trial solution $\hat{\psi}_{t}$ can be constructed by adding its individual parts given in Eqs.~(\ref{Eq:44}) and (\ref{Eq:46}) as follows
\begin{equation}
    \hat{\psi}_{t} = \hat{\psi}_{t,I}+\hat{\psi}_{t,II}.
\end{equation}
Since $\hat{\psi}_{t}$ satisfies the boundary conditions prior to the training process, the corresponding loss function $\mathcal{L}_b$ (Eq.~(\ref{Eq:17})) can be dropped, and the optimization problem in Eq.~(\ref{Eq:6}) can be written as
\begin{equation}
    \min_{\theta} \quad \mathcal{L}(\mathbf{x};\theta), \quad \mathbf{x}\in \{\Omega \cup \partial\Omega\}
\end{equation}
where
\begin{equation}
        \mathcal{L}(\mathbf{x};\theta) = \frac{1}{N}\sum_{i=1}^{N}\left\|\nabla^2\hat{\psi}_t(\mathbf{x};\theta)+k^2\hat{\psi}_t(\mathbf{x};\theta)\right\|^2_2.
\end{equation}
Here, $N$ is the total number of collocation points within the domain, including the boundary points. For the rectangular domain considered, $\mathcal{L}$ can be written as
\begin{multline}
    \mathcal{L}(x,y;\theta) = \frac{1}{N}\sum_{i=1}^{N}\left\|\left.\left(\frac{\partial^2}{\partial x^2}\hat{\psi}_t(x,y^{(i)};\theta)\right)\right|_{x=x^{(i)}}\right. \\
    +\left.\left(\frac{\partial^2}{\partial y^2}\hat{\psi}_t(x^{(i)},y;\theta)\right)\right|_{y=y^{(i)}} \\
    +\left.k^2\hat{\psi}_t(x^{(i)},y^{(i)};\theta)\right\|^2_2. \label{Eq:53}
\end{multline} \vspace{-15pt}
	\begin{figure}[h!]
	\includegraphics[scale=0.85]{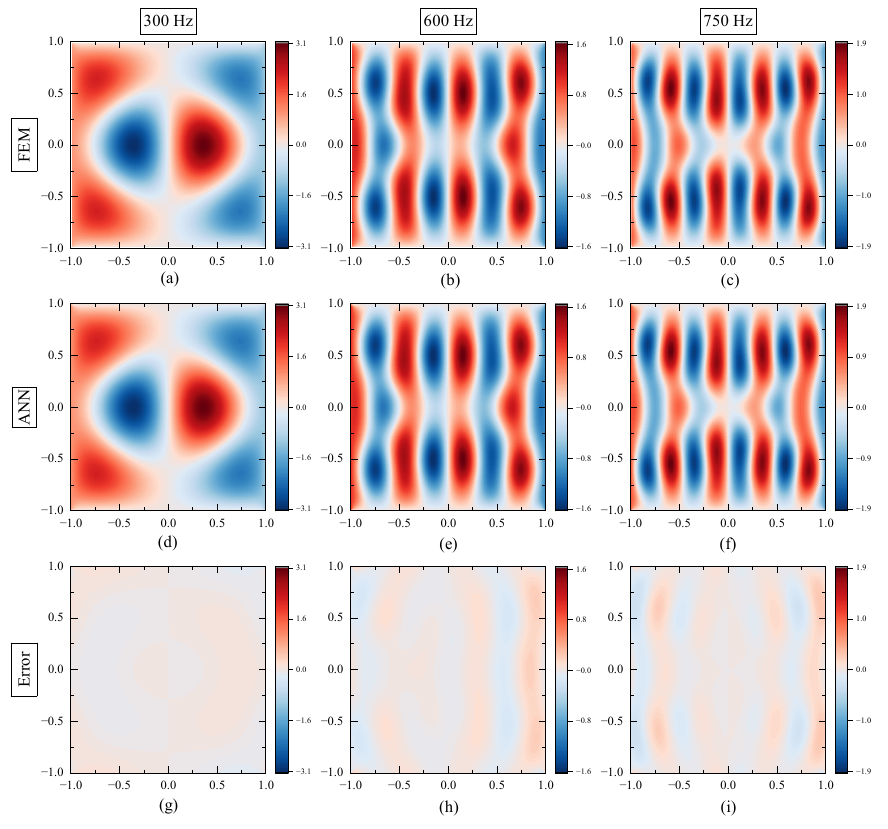}\centering
	\caption{\label{fig:14}{Acoustic field distribution at 300 Hz, 600 Hz, and 750 Hz: (a, b, c) from FEM, (d, e, f) from ANN (with the trial solution method), and (g, h, i) are the errors (FEM-ANN).}}
	\end{figure}
 
Fig.~\ref{fig:14} shows the acoustic field distribution obtained from the ANN using the trial solution method against that obtained from the FEM. Here, ANN results were obtained using the network architecture mentioned in Table~\ref{tab:1}. Earlier, it was reported in Section~\ref{Sec:3} that the equal weighting method, that is, $\mathcal{L}=\mathcal{L}_d+\mathcal{L}_b$ is able to predict the acoustic field up to 600 Hz, and fails at 750 Hz (Fig.~\ref{fig:4}). It can be observed that the trial solution is able to circumvent the \emph{vanishing gradient problem} and is able to predict the acoustic field accurately at 300 Hz, 600 Hz, and 750 Hz, which can be confirmed by comparing the error plots in the bottom row of Fig.~\ref{fig:14}.
 	\begin{figure}[h!]
	\includegraphics[scale=0.9]{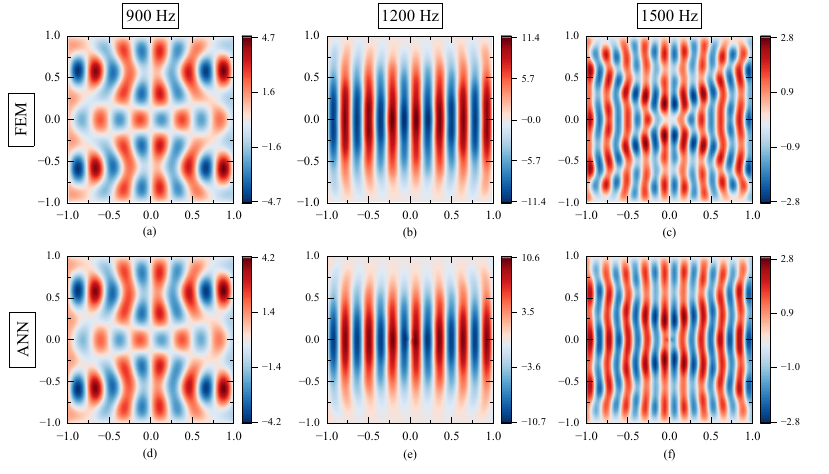}\centering
	\caption{\label{fig:15}{Acoustic field distribution at 900 Hz, 1200 Hz, and 1500 Hz: (a, b, c) from FEM, (d, e, f) from ANN (with the trial solution method).}}
	\end{figure}
 
 Fig.~\ref{fig:15} shows the extension of the methodology up to 1500 Hz starting from 900 Hz, in steps of 300 Hz. It can be observed that the trial solution method is able to predict the acoustic field with reasonable accuracy up to the frequency considered. These results confirm the superiority of the trial solution method over the Lagrange multiplier method, which fails to predict the acoustic field at 900 Hz as demonstrated in Fig.~\ref{fig:7}. It should be noted here that the error plots are excluded here and in the subsequent sections because the deviation of the predicted results from those of the FEM is marginal. The applicability of the proposed methodology to the circular and elliptical domains is presented in the following sections.

\subsection{Acoustic field in a circular domain} \label{Sec:4.3}
\begin{figure}[h!]
	\includegraphics[scale=0.9]{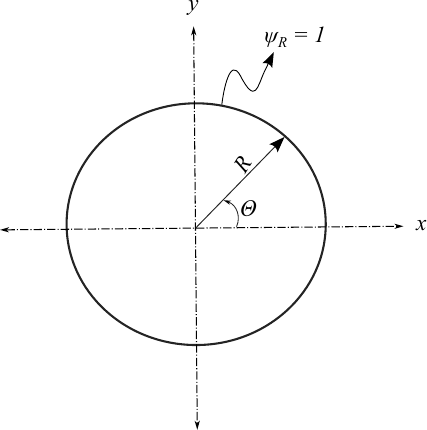}\centering
	\caption{\label{fig:16}{Schematic diagram of a circular domain with the boundary condition.}}
	\end{figure}

 Let us consider a circle of radius $R$ as shown in Fig.~\ref{fig:16} with a unit acoustic field on the boundary. Since the circular domain has one continuous boundary, it will have only one distance function $\phi$ which is equal to the equivalent distance function $\phi_e$. 

 Using the theory of R-functions, the equivalent distance function $\phi_e$ that vanishes on the boundary and is positive inside the domain can be constructed from the geometric representation of the circle in the cartesian coordinate system as \cite{Rvachev1995}
\begin{equation}
    \phi_e(\mathbf{x}) = \frac{R^2-\left[(x-x_c)^2+(y-y_c)^2\right]}{2R}, \label{Eq:54}
\end{equation}
where
\begin{align}
    x &= r\cos\Theta, \label{Eq:55}\\
    y &= r\sin\Theta. \label{Eq:56}
\end{align}
Here, $r\in [0, R]$ and $\Theta\in [0, 2\pi]$. Figure~17a shows the equivalent distance function $\phi_e$ of the circle of a unit radius, centered at the origin $(0, 0)$. It can be observed that $\phi_e$ vanishes on the boundary and is positive inside the domain. From $\phi_e$, the weight function $W(\mathbf{x})$ can be obtained as follows
\begin{equation}
    W(\mathbf{x}) = 1-\phi_e, \label{Eq:58}
\end{equation}
and is shown in Fig.~17b. Note here that $W(\mathbf{x})$ can be found from Eq.~(\ref{Eq:58}) for any circle of arbitrary radius due to the fact that $\phi_e$ is normalized with respect to the diameter of the circle in Eq.~(\ref{Eq:54}). 
 	\begin{figure}[h!]
	\includegraphics[scale=1.8]{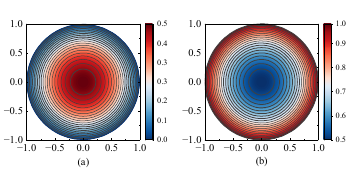}\centering
	\caption{\label{fig:17}{Equivalent distance function ($\phi_e$) and weight ($W$) for the circular domain: (a) $\phi_e$, (b) $W$.}}
	\end{figure}
 
\begin{figure}[h!]
	\includegraphics[scale=0.9]{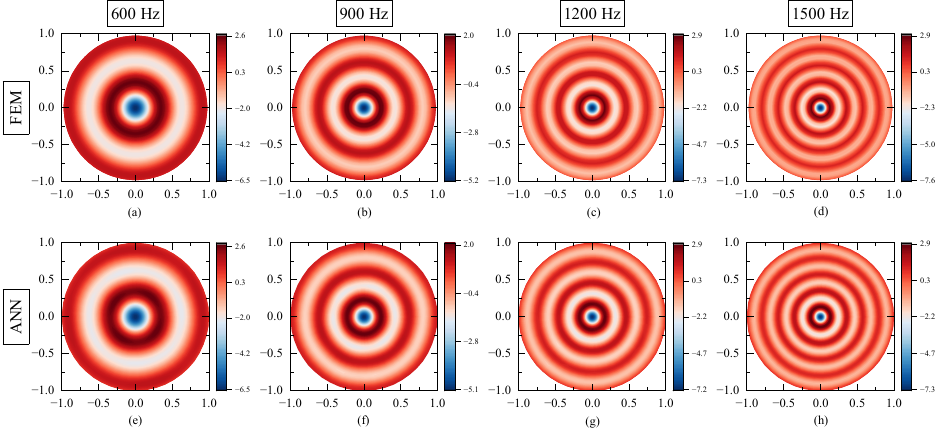}\centering
	\caption{\label{fig:18}{Acoustic field distribution at 600 Hz, 900 Hz, 1200 Hz, and 1500 Hz: (a, b, c, d) from FEM, (e, f, g, h) from ANN (with the trial solution method).}}
	\end{figure}

Now, the trial solution $\hat{\psi}_t$ can be constructed as 
\begin{equation}
    \hat{\psi}_t(x,y;\theta) = W\psi_R+\phi_e\hat{\psi}(x,y;\theta).
\end{equation}
This can be substituted in Eq.~(\ref{Eq:53}) and the loss function $\mathcal{L}$ can be evaluated. For this purpose, the circle of unit radius is divided into 10000 random collocation points ($N=$ 10000). Using the network architecture mentioned in Table~\ref{tab:1}, a neural network is constructed and the loss function is minimized. The acoustic field thus obtained from 600 Hz to 1500 Hz in steps of 300 Hz is shown in Fig.~\ref{fig:18} against that obtained from the FEM. It can be observed that there is good agreement between the proposed method and the FEM at all frequencies considered. Note here that as prediction of the acoustic field using ANN at lower frequencies is not critical, comparison of the acoustic field distribution at 300 Hz is excluded from the results. 

Note here that the Helmholtz equation has been solved in the Cartesian coordinate system instead of the polar coordinate system using the transformation Eqs.~(\ref{Eq:55}) and (\ref{Eq:56}). Alternately, one can substitute these transformation equations into the Laplacian $\nabla^2$ of Eq.~(\ref{Eq:1}), and the resulting Bessel equation \cite{Bowman2003}
\begin{equation}
    \left[\frac{1}{r}\frac{\partial}{\partial r}\left(r\frac{\partial}{\partial r}\right)+\frac{1}{r^2}\frac{\partial^2}{\partial\Theta^2}+k^2\right]\psi(r,\Theta) = 0
\end{equation}
can be solved using the proposed formulation. In this work, the former method has been used due to its simplicity.

\subsection{Acoustic field in an elliptical domain} \label{Sec:4.4}
\begin{figure}[h!]
	\includegraphics[scale=1]{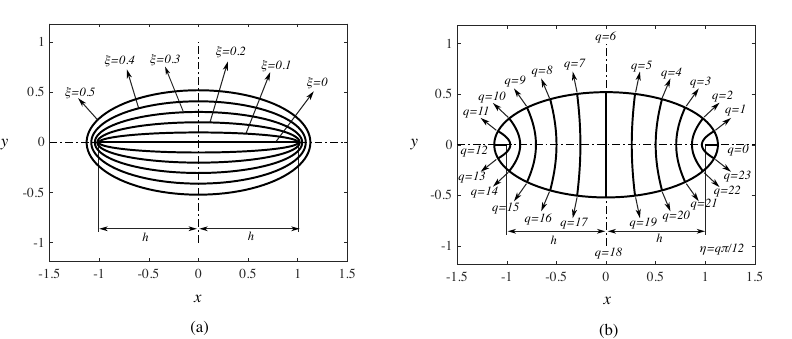}\centering
	\caption{\label{fig:19}{Parametric representation of family of ellipses and a hyperbolae: (a) Family of ellipses, (b) Family of hyperbolae.}}
	\end{figure}

It is known that the equation of an ellipse with $a$ and $b$ as the semi-major and semi-minor axes, respectively, can be written in the Cartesian coordinate system as follows \cite{McLachlan1947}
\begin{equation}
    \frac{x^2}{a^2}+\frac{y^2}{b^2} = 1. \label{Eq:60}
\end{equation}
Let $a=l\cosh\xi$ and $b=l\sinh\xi$, then Eq.~(\ref{Eq:60}) becomes
\begin{equation}
    \frac{x^2}{l^2\cosh^2\xi}+\frac{y^2}{l^2\sinh^2\xi} = 1, \label{Eq:61}
\end{equation}
and represents a family of confocal ellipses with the focal distance $l=\sqrt{a^2-b^2}$. Here, $\xi$ represents the radial coordinate whose value is zero along the line of foci and takes some positive value at each ring as shown in Fig.~\ref{fig:19}a.

Similarly, 
\begin{equation}
    \frac{x^2}{l^2\cos^2\eta}-\frac{y^2}{l^2\sin^2\eta} = 1, \label{Eq:62}
\end{equation}
represents a family of hyperbolae whose semi-conjugate axis is $a=l\cos\eta$, semi-transverse axis $b=l\sin\eta$, and the focal distance $l=\sqrt{a^2+b^2}$. Here, $\eta$ represents the angular coordinate that varies from 0 to $2\pi$ while passing each quadrant as shown in Fig.~\ref{fig:19}b.

\begin{figure}[h!]
	\includegraphics[scale=0.7]{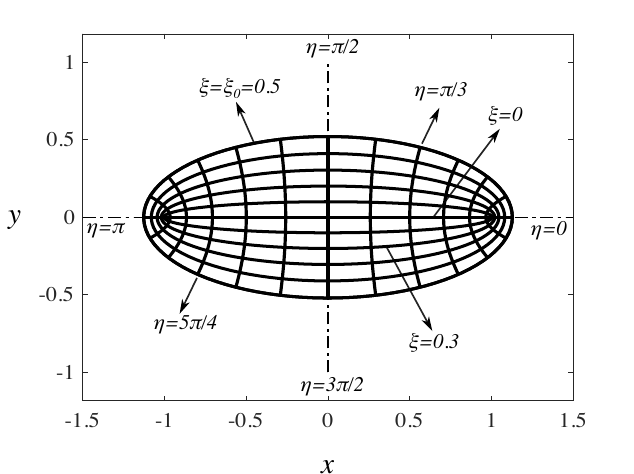}\centering
	\caption{\label{fig:20}{Elliptical coordinate system.}}
	\end{figure}

The two conics given in Eqs.~(\ref{Eq:61}) and (\ref{Eq:62}) intersect orthogonally at \cite{McLachlan1947} 
\begin{align}
    x &= l\cosh\xi\cos\eta, \label{Eq:63} \\
    y &= l\sinh\xi\sin\eta, \label{Eq:64}
\end{align}
and forms an elliptical coordinate system $(\xi,\eta)$, where $\xi\in[0,\xi_0]$ and $\eta\in[0,2\pi]$, as shown in Fig.~\ref{fig:20}. 
 
Since the elliptical domain is also having one continuous boundary, its equivalent distance function $\phi_e$ and the weight function $W$ can be constructed, respectively, as follows 
\begin{align}
    \phi_e(\mathbf{x}) &= \frac{1-\left[(x^2/a^2)+(y^2/b^2)\right]}{2}, \\
    W(\mathbf{x}) &= 1-\phi_e.
\end{align} 
	\begin{figure}[h!]
	\includegraphics[scale=2.3]{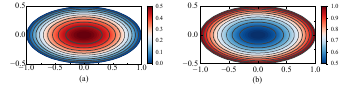}\centering
	\caption{\label{fig:21}{Equivalent distance function ($\phi_e$) and weight ($W$) for the elliptical domain: (a) $\phi_e$, (b) $W$.}}
	\end{figure}

Fig.~\ref{fig:21} shows $\phi_e$ and $W$ for an ellipse of the semi-major axis $a=$ 1 m and the semi-minor axis $b=$ 0.5 m. It can be observed that, as in any other case, $\phi_e$ vanishes on the boundary and is positive inside the domain.
 	\begin{figure}[h!]
	\includegraphics[scale=1.05]{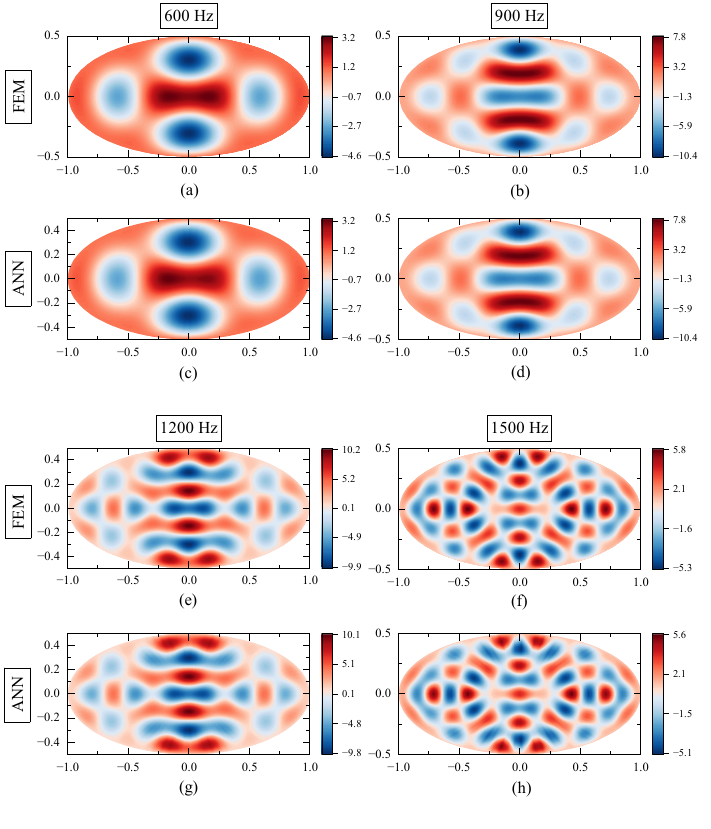}\centering
	\caption{\label{fig:22}{Acoustic field distribution at 600 Hz, 900 Hz, 1200 Hz, and 1500 Hz: (a, b, e, f) from FEM, (c, d, g, h) from ANN (with the trial solution method).}}
	\end{figure}

Substituting Eqs.~(\ref{Eq:63}) and (\ref{Eq:64}) in Eq.~(\ref{Eq:1}) yields the Helmholtz equation in the elliptical coordinate system as follows \cite{McLachlan1947,Munjal2014}
\begin{equation}
    \left[\frac{2}{l^2\left(\cosh 2\xi-\cos 2\eta\right)}\left(\frac{\partial^2}{\partial\xi^2}+\frac{\partial^2}{\partial\eta^2}\right)+k^2\right]\psi(\xi, \eta) = 0. 
\end{equation}
For the given values of $\psi(\xi_0,\eta)$ at the boundary, the above equation can be solved using the proposed methodology. Fig.~\ref{fig:22} shows the acoustic field for an elliptical geometry whose semi-major axis $a=$ 1 m and semi-minor axis $b=$ 0.5 m, that is, $h=$ 0.866, $\xi=\xi_0=$ 0.549, and $\eta\in[0,2\pi]$, at frequencies ranging from 600 Hz to 1500 Hz, in steps of 300 Hz. Here, the chosen elliptical domain is divided into 10000 random collocation points ($N=$ 10000) using the Eqs.~(\ref{Eq:63}) and (\ref{Eq:64}). The neural network architecture mentioned in Table~\ref{tab:1} has been used to optimize the loss function. It can be observed from Fig.~\ref{fig:22} that the acoustic field obtained from the neural network formulation is in good agreement with that obtained from the FEM. The results also indicate the ability of the proposed formulation to predict acoustic fields with relatively high modal density, which can be evidenced from the predictions at 1200 Hz and 1500 Hz.

 \subsection{Handling discontinuities at the corners}
\begin{figure}[h!]
	\includegraphics[scale=1]{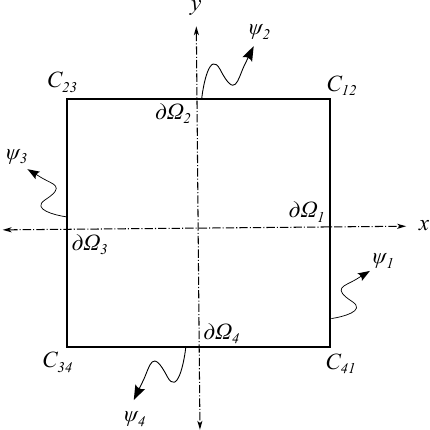}\centering
	\caption{\label{fig:23}{Schematic diagram of the rectangular domain with four boundaries $\partial\Omega_i$.}}
    \end{figure}

 Both circular and elliptical domains have continuous boundaries. It indicates that one distance function is sufficient to construct the equivalent distance function ($\phi_e$) and/or the inverse distance weight function(s) ($W$ or $W_i$). In contrast, the rectangular geometry is composed of four boundaries $\partial\Omega_i$ with respective boundary values $\psi_i$, as shown in Fig.~\ref{fig:23}. Here, the subscript $i=$ 1, 2, 3, and 4 denotes the indices of the boundaries. 
 
 It can be seen that the boundary values are discontinuous at the corners $C_{12}$, $C_{23}$, $C_{34}$, and $C_{41}$. In addition, the inverse distance functions $W_i$ also become discontinuous at the same corners. The handling of these discontinuities is essential for the proper estimation of the acoustic field in the rectangular domains. 

\subsubsection{Handling discontinuities in $W_i$}
\begin{figure}[h!]
	\includegraphics[scale=1.7]{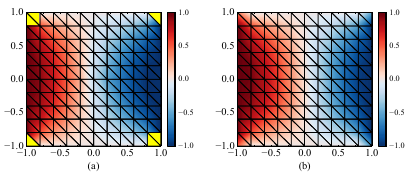}\centering
	\caption{\label{fig:24}{Effect of discontinuities in $W_i$ at the corners of the rectangular domain on the acoustic field: (a) without $\epsilon$-approximation, (b) with $\epsilon$-approximation.}}
    \end{figure}

The inverse distance functions ($W_i$) become discontinuous at the corner points and introduces errors (NaN) while predicting the acoustic field. For instance, consider the corner $C_{12}$ which is a common point to the boundaries $\partial\Omega_1$ and $\partial\Omega_2$. According to Eq.~(\ref{Eq:45}), the inverse distance function $W_1$ can be calculated as
\begin{equation}
    W_1(\mathbf{x}) = \frac{\phi_2\phi_3\phi_4}{\phi_2\phi_3\phi_4+\phi_1\phi_3\phi_4+\phi_1\phi_2\phi_4+\phi_1\phi_2\phi_3}. 
\end{equation}
Ideally, 
\begin{equation}
    W_1(\mathbf{x}) = 1, \quad \forall \, \mathbf{x}\in\partial\Omega_1
\end{equation}
as 
\begin{equation}
    \phi_1(\mathbf{x}) = 0, \quad \forall \, \mathbf{x}\in\partial\Omega_1.
\end{equation}
However, the corner point $C_{12}$ is common to both boundaries, that is, $C_{12}\in\partial\Omega_1$, and $C_{12}\in\partial\Omega_2$. It implies that
\begin{equation}
    \phi_1(\mathbf{x}) = \phi_2(\mathbf{x}) = 0, \quad \text{at} \quad \mathbf{x}=C_{12}, \label{Eq:71}
\end{equation}
and $W_1(\mathbf{x})$ becomes undefined at $\mathbf{x}=C_{12}$. This problem also exists at other three corner points. As the inverse distance weight function becomes numerically NaN, the acoustic field cannot be predicted at these corners. This phenomenon is demonstrated in Fig.~\ref{fig:24}a. The yellow color elements represent the undefined acoustic field at the corners. Here, coarse grid is used for better visualization purposes. 
    
The authors realized that the problem of discontinuity in $W_i$ at the corner can be bypassed by adding or subtracting an infinitely smaller value $\epsilon$ from the coordinates of the corner points, that is, $C_{12}=(x_{12}, y_{12})\approx (x_{12}\pm\epsilon, y_{12}\pm\epsilon)$ while calculating $W_i$. This approximation apparently shifts the corner points either inside or outside the domain by a distance $\sqrt{2}\,\epsilon$, and eliminates the situation occurring in Eq.~(\ref{Eq:71}). The error occurring in the estimation of the distance function $\phi_1$ at the corner point due to this approximation can be calculated from Eqs.~(\ref{Eq:33}), (\ref{Eq:34}), and (\ref{Eq:35}). For the considered rectangular geometry, it is found that 
\begin{equation}
    \delta\phi_1 = \epsilon\left(1-\frac{\epsilon}{8}\right).  
\end{equation}
In this work, $\epsilon$ is taken in the order of $\mathcal{O}(10^{-17})$. Therefore, $\delta\phi_1$ can be neglected. The same is also valid for other corner points. Figure~\ref{fig:24}b shows the acoustic field after eliminating the boundary discontinuities at all four corners using $\epsilon$-approximation.  
    \begin{figure}[h!]
	\includegraphics[scale=1.6]{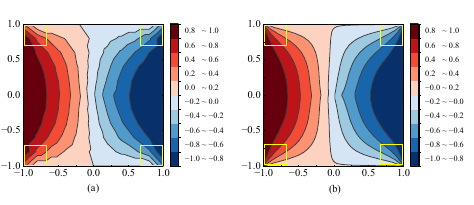}\centering
	\caption{\label{fig:25}{Boundary-value discontinuities at the corners of the rectangular domain: (a) Coarse grid (10 $\times$ 10), (b) Fine grid (80 $\times$ 80).}}
    \end{figure}
    
\subsubsection{Handling boundary-value discontinuities}
Let us consider the corner $C_{12}$ again, which connects the boundaries $\partial\Omega_1$ and $\partial\Omega_2$. The first part of the trial solution corresponding to these two boundaries can be written as
\begin{equation}
    \hat{\psi}_{t,I}(\mathbf{x}) = W_1(\mathbf{x})\psi_1+W_2(\mathbf{x})\psi_2,
\end{equation}
 where $W_1$ and $W_2$ are the inverse distance functions associated with the boundaries $\partial\Omega_1$ and $\partial\Omega_2$, respectively. 

Consider a point $\mathbf{x}\in\partial\Omega_1$, and estimate the value of $\hat{\psi}_{t,I}$ as $\mathbf{x}$ approaches $\partial\Omega_2$, that is, 
\begin{equation}
    \lim_{\mathbf{x}\in\partial\Omega_1;\,\mathbf{x}\to\partial\Omega_2} \hat{\psi}_{t,I}(\mathbf{x}) = \lim_{\mathbf{x}\in\partial\Omega_1;\,\mathbf{x}\to\partial\Omega_2} W_1(\mathbf{x})\psi_1+\lim_{\mathbf{x}\in\partial\Omega_1;\,\mathbf{x}\to\partial\Omega_2} W_2(\mathbf{x})\psi_2,
\end{equation}
 \begin{equation}
   \implies \lim_{\mathbf{x}\in\partial\Omega_1;\,\mathbf{x}\to\partial\Omega_2} \hat{\psi}_{t,I}(\mathbf{x}) = \psi_1. \label{Eq:75}
\end{equation}
Similarly, consider a case where $\mathbf{x}\in\partial\Omega_2$ and $\mathbf{x}\to\partial\Omega_1$, that is, 
\begin{equation}
    \lim_{\mathbf{x}\in\partial\Omega_2;\,\mathbf{x}\to\partial\Omega_1} \hat{\psi}_{t,I}(\mathbf{x}) = \lim_{\mathbf{x}\in\partial\Omega_2;\,\mathbf{x}\to\partial\Omega_1} W_1(\mathbf{x})\psi_1+\lim_{\mathbf{x}\in\partial\Omega_2;\,\mathbf{x}\to\partial\Omega_1} W_2(\mathbf{x})\psi_2,
\end{equation}
 \begin{equation}
   \implies \lim_{\mathbf{x}\in\partial\Omega_2;\,\mathbf{x}\to\partial\Omega_1} \hat{\psi}_{t,I}(\mathbf{x}) = \psi_2. \label{Eq:77}
\end{equation}
Eqs.~(\ref{Eq:75}) and (\ref{Eq:77}) indicates that $\hat{\psi}_{t,I}(\mathbf{x})$ is discontinuous at $\mathbf{x}=C_{12}$. This phenomenon can be seen in Fig.~\ref{fig:25}a, and its prevalence can also be observed at other corners. Fortunately, these discontinuities are local phenomena and can be limited to the corners through the \emph{finer} domain discretization as demonstrated in Fig.~\ref{fig:25}b \cite{Rvachev2001}. 

\section{Conclusion} \label{Sec:5}
A neural network formulation is presented to solve the two-dimensional Helmholtz equation. Two popular methods known as \emph{Lagrange multiplier method} and \emph{trial solution method} that convert the constrained optimization problem into an unconstrained optimization problem are demonstrated. It is observed that the former method encounters a vanishing gradient problem as the frequency of analysis increases. Initially, automatic $\lambda$ update algorithms seem to mitigate this problem. However, the analysis reveals that their applicability is limited by the manual tuning of the hyperparameters. Therefore, the latter method, which is used by researchers to solve other governing equations, has been adopted here to solve the Helmholtz equation. 

Through its elegant construction, the trial solution method is able to bypass the vanishing gradient problem by eliminating the loss function associated with the boundary conditions from the optimization process. By its form, the method appears to be obvious and trivial to implement. However, its construction in two-dimensional settings is challenging and requires knowledge from interdisciplinary concepts such as transfinite interpolation, inverse distance weighting interpolation, implicit functions, R-functions, set theory, etc. The complexity in the formulation arising from the amalgamation of these concepts can be justified by its capability to predict the results at par with the well-established finite element methods. In this work, the analysis is limited to rectangular, circular, and elliptical domains. However, one can extend this formulation to the domains of other shapes as well. In addition, a single neural network architecture has been used to predict the acoustic field in all three domain configurations. Formulations that exhibit such a nature will have an ability to emerge as a neural network-based solvers in the near future. 

\section*{Data availability}
Data will be made available on request.

\section*{Acknowledgments}
The authors acknowledge the support received from the Department of Science and Technology, and Science and Engineering Research Board (SERB), Government of India towards this research.

\appendix
\setcounter{figure}{0}
\setcounter{table}{0}
\section{2-D FEM modeling of a rectangular domain} \label{Append:A}
The 2-D FEM analysis to find the acoustic field distribution in a given rectangular domain involves three steps: 1) discretization of the domain, 2) application of the material properties and boundary conditions, and 3) post-processing. These steps are described here.
   \begin{figure}[h!]
	\includegraphics[scale=0.8]{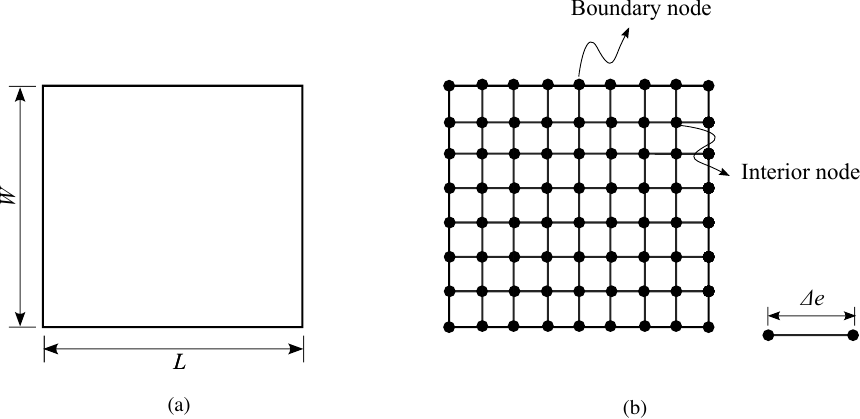}\centering
	\caption{\label{fig:A1}{Schematic diagram of a rectangular domain with discretization.}}
    \end{figure}

Consider a rectangular domain of cross-sectional $L\times W$ m$^2$ as shown in Fig.~\ref{fig:A1}a. In order to perform the FEM analysis, the domain has to be discretized into a number of elements as shown in Fig.~\ref{fig:A1}b. Each element is connected to its adjacent element through a common node. The maximum size of each element should be considered in such a way that the discretization captures the acoustic field at the minimum wavelength present in the analysis. This is possible by choosing the maximum element size $\Delta e_{\textup{max}}$ always less than one-sixth of the minimum wavelength ($\lambda_{\textup{min}}$) present in the analysis, that is,
\begin{equation}
    \Delta e_{\textup{max}} \leq \frac{\lambda_{\textup{min}}}{6} = \frac{c}{6f_{\textup{max}}},
\end{equation}
where $c$ is the speed of sound, and $f_{\textup{max}}$ is the maximum frequency of the analysis.

It is known that sound propagation requires the medium to have inertia and elasticity. Hence, these properties will be specified in terms of the density of the medium ($\rho$) and the speed of sound in it ($c$). These properties will be assigned to all the elements of the domain. The boundary conditions will be applied to the boundary nodes of the domain. 

Now, the model is solved at discrete frequencies in the COMSOL Multiphysics solver \cite{Comsol2022}. The acoustic field distribution at the required frequencies is extracted for validation purposes. By following the same procedure, the acoustic field distribution in the circular and elliptical domains can be obtained. In this work, square elements have been used for the rectangular domain, whereas triangular elements have been used for the circular and elliptical domains with first-order interpolation. 

\section{Parameters of the Gaussian distributions} \label{Append:B}
The Gaussian fits for the gradients of $\mathcal{L}_d$ and $\mathcal{L}_b$ are calculated using the following equation in OriginLab \cite{Origin2023}
\begin{equation}
    y(x) = y_0+\frac{A}{w\times\sqrt{\frac{\pi}{4\ln(2)}}}\,\textup{e}^{-\frac{4\ln(2) (x-x_c)^2}{w^2}}.
\end{equation}
The values of the parameters for $\nabla\mathcal{L}_d$ and $\nabla\mathcal{L}_b$ are tabulated below
\begin{table}[h!]
    \centering
    \begin{tabular}{@{}clc@{}}
        \toprule
        Parameters & \hfil $\nabla\mathcal{L}_d$ & $\nabla\mathcal{L}_b$ \\
        \midrule
        $y_0$ & 175.652$\,\pm\,$28.293 & 44.769$\,\pm\,$19.526 \\
        $x_c$ & $-$0.112$\,\pm\,$0.386 & 2.163$\times$10$^{-4}\,\pm\,$9.751$\times$10$^{-4}$ \\
        $A$ & 20844.832$\,\pm\,$1521.778 & 364.824$\,\pm\,$9.208 \\
        $w$ & 12.798$\,\pm\,$0.9971 & 0.094$\,\pm\,$0.003 \\
        \bottomrule
    \end{tabular}
    \caption{Gaussian parameters for $\nabla\mathcal{L}_d$ and $\nabla\mathcal{L}_b$.}
    \label{tab:2}
\end{table}
\newpage
\bibliographystyle{elsarticle-num} 
\bibliography{references}

\end{document}